# Random Occlusion-recovery for Person Re-identification

Di Wu[1], Kun Zhang[1], Fei Cheng[4], Yang Zhao[4], Qi Liu[2*], Chang-An Yuan[3*] and De-Shuang Huang[1*]

[1]Institute of Machine Learning and Systems Biology, School of Electronics and Information Engineering, Tongji University

[2]School of Computing, Edinburgh Napier University, 10 Colinton Road, Edinburgh, EH10 5DT UK

[3]Science Computing and Intelligent Information Processing of Guang Xi Higher Education Key Laboratory, Guangxi Teachers Education University, Nanning, Guangxi, 530001, China

[4]Beijing E-Hualu Info Technology Co., Ltd Beijing, China

*Abstract.* As a basic task of multi-camera surveillance system, person re-identification aims to re-identify a query pedestrian observed from non-overlapping multiple cameras or across different time with a single camera. Recently, deep learning-based person re-identification models have achieved great success in many benchmarks. However, these supervised models require a large amount of labeled image data, and the process of manual labeling spends much manpower and time. In this study, we introduce a method to automatically synthesize labeled person images and adopt them to increase the sample number per identity for person re-identification datasets. To be specific, we use block rectangles to randomly occlude pedestrian images. Then, a generative adversarial network (GAN) model is proposed to use paired occluded and original images to synthesize the de-occluded images that similar but not identical to the original image. Afterwards, we annotate the de-occluded images with the same labels of their corresponding raw images and use them to augment the number of samples per identity. Finally, we use the augmented datasets to train baseline model. The experiment results on CUHK03, Market-1501 and DukeMTMC-reID datasets show that the effectiveness of the proposed method.

*Keywords*—Generative Adversarial Network, Person Re-identification.

## 1. Introduction

Person re-identification (ReID) is an important task in many computer vision systems, such as behavioral understanding, threat detection and video surveillance. Given a query person image, it aims to re-identify the person observed by non-overlapping multiple cameras or across different time with a single camera. The task has drawn significant attention in computer vision community. So far, it is still a challenge issue for the appearance of a person may suffer dramatic change under different camera views. Traditional hand-craft methods address the person ReID issue through either finding discriminative feature representations [1-4] or exploiting a suitable distance metric function [5-7]. When feature representations are obtained, a distance metric function is applied to estimate whether the paired inputs are the same pedestrian or not. Recently, enlightened by the success of deep learning technology, a large number of works introduce the technology to address the person ReID issue and achieve many promising performances. Most of recent state-of-the-art person ReID models are based on deep learning technology. Both the training processes of the models require a large amount of labeled data. However, existing available public person ReID datasets are limited in their scales, especially for the number of images per identity. For example, the average numbers of images per identity for large-scale ReID datasets like CUHK03 [8], Market-1501 [9] and DukeMTMC-ReID [10] are only 9.6, 17.2 and 23.5, respectively. Using such scale datasets to train the deep models may lead to over-fitting issue and affect the robustness of them. Moreover, rely on human annotating is expensive and time-consuming, for one not only needs to draw a bounding box for pedestrians, but also needs to assign each of them an ID. Recently, several GAN-based data augmentation approaches [11, 12] have been introduced to alleviate this issue. However, due to the appearances and backgrounds of the generated samples are far away from their original identities, most of approaches above have to assign the generated sample a new label. Therefore, the number of images per identity cannot be increased. In order to solve the above problem, our motivation is to generate samples that similar enough but not identical to their original images and to annotate them with their original label information, thus the number of images per identity can be increased.

To this end, as shown in Figure 2, firstly, we use block rectangles to randomly occlude the original ReID training images. Secondly, we use the paired occluded images and their ground-truths to train a de-occluding GAN model. Then the trained GAN model is used to generate the de-occluded images with the same labels of their corresponding ground-truths. Finally, these generated images are combined with the original ReID training images to train the baseline model together.

In summary, we make the following contributions:
(1) We introduce a data augmentation method that can generate person images which similar enough but not identical to the original pedestrian images;
(2) We attempt to assign the original annotation information to the generated pedestrian images, thus increasing the sample number per identity;
(3) We show that the proposed data augmentation method with original annotation information improved the person ReID accuracy with the baseline model.

## 2. Related work

In this section, we briefly review the existing literature for CNN, LSTM and GAN-based methods in person ReID community.

***CNN/LSTM-Based Person ReID.*** A large number of deep learning technology-based models are proposed to address the person ReID task. Some of them regard the task as a classification issue. In order to complement the CNN features, Wu et al. [13] propose a fusion feature network (FFN) which can incorporate a variety of hand-crafted features into deep features. Xiao et al. [14] combine multiple datasets to train the CNN model and propose a domain guide dropout strategy to discard worthless neurons for each domain dataset to keep the model in the right track. Lin et al. [15] hold that the person ID recognition learns global representations while attribute classification extracts local aspects (i.e., age, gender, bag and so on). Considering the difference and similarity of the two tasks, they propose combining the attribute classification and identification losses to focus on the local and global aspects of a pedestrian image jointly. It is worth mentioning that distance metric-based deep models are also frequently adopted in person ReID community. Ding et al. [16] first introduce the triplet model to address the ReID task. Cheng et al. [17] propose an improved triplet loss which can push the different person images farther from each other, and simultaneously pull the same person images closer under the learned deep feature space. Chen et al. [18] introduce a quadruplet loss to enhance the generalization ability of the deep model. There are also exist some studies adopt LSTM model to learn the spatial information or learn attention-based features for person images. A Siamese LSTM model is proposed by [19] for person

ReID. The authors initially divide the image into several rigid parts, and then they extract hand-craft feature like local maximal occurrence and SILTP for each part. Finally, they use a LSTM model to leverage the contextual information of the local descriptors. Liu et al. [20] propose a LSTM-based attention model that can dynamically produce part attention features by a recurrent way.

***Data augmentation-based deep model.*** Recently, in order to solve the data scarcity problem, some works attempt to use the GAN model to extend the scale of person ReID datasets. Zheng et al. [11] propose a generative adversarial network to generate unlabeled samples for ReID datasets. They introduce a label smoothing regularization for outliers (LSRO) method to annotate the generated samples. To reduce the domain gap between different datasets, Wei et al. [21] introduce a transfer generative adversarial network which alleviates the domain gap through transferring persons in *A* dataset to *B* dataset. The transferred pedestrians from *A* keep the similar styles such as backgrounds, lightings with dataset *B*. Similarly, Zhong et al. [12] propose a camera style adaption model to transfer the style of images captured by one camera to another. They adopt label smoothing regularization (LSR) to mitigate the influence of noise introduced by the style-transferred samples. In [22], a pose-normalization GAN model is introduced to reduce the influence of pose variation. Our approach differs from these existing data augmentation-based models mentioned above. Specifically, the works [11] and [12] assign a new annotation information for the generated samples. This is because the qualities of generated samples of them are too low and the appearances of the generated samples are far away from their corresponding raw images. Thus, labeling them with original IDs information as training data can negatively affect the training process of deep model. However, retaining the original annotation information for the generated samples is important, especially for the datasets with limited samples per identity, because the number per identity can be increased only when the original label is preserved. The studies [21] and [22] need to introduce additional mask operation and pose estimation, respectively. Furthermore, both the two works differ from ours in methodology and motivation. In our method, we directly use the designed GAN to generate samples that similar enough but not exactly same to their ground-truths without any extra operation. The generated samples are annotated with original label information, and hence the number of image samples per identity is enlarged.

## 3. The proposed method

In this section, we first present the random occlusion method. Then, we illustrate the architecture of the proposed GAN model. Finally, the loss functions adopted by this model are discussed.

### 3.1 Random Occlusion

In the training phase, we add a random occlusion for each training image. As shown in Figure 1, a rectangle region of the image is randomly selected for erasing by an occlusion. The occlusion can be a black box or a white box with 0, 255 values for its pixels, respectively. In this paper, the values of pixels on R, G and B channels for the occlusion are set to the mean pixel values of R, G and B channels of the ReID dataset, respectively. For example, for the Market-1501 dataset, we set the pixel values on R, G and B channels for the occlusion to 105.3, 99.6 and 97.9, respectively.

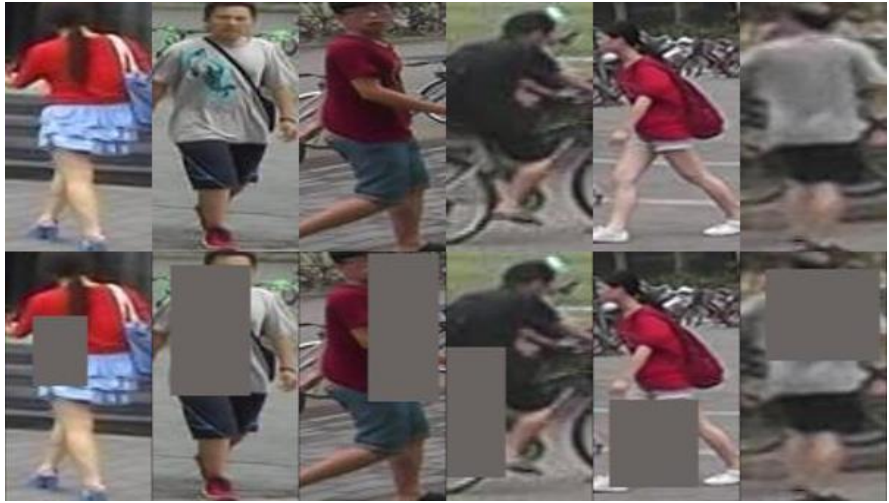

**Figure 1**. The occluded samples

### 3.2 Network architecture

In this study, the proposed model consists of three major parts, i.e. generator, discriminator and loss function. Similar to the original GAN [23], the network contains two sub-networks: a generator sub-network *G* and a discriminator sub-network *D*. While training, we put the paired raw and occluded training images into the generator *G*. The primary goal of the *G* network is to synthesize a de-occluded image as close as possible to the raw training image to fool the discriminator *D*. The sub-network *D* aims to distinguish the fake de-occluded image generated by *G* from its corresponding real image. In other words, the discriminator *D* serves as a supervisor to improve the quality of the synthesized samples. In this section, we first discuss the framework of the proposed network, and then detailedly introduce the loss functions used in the network.

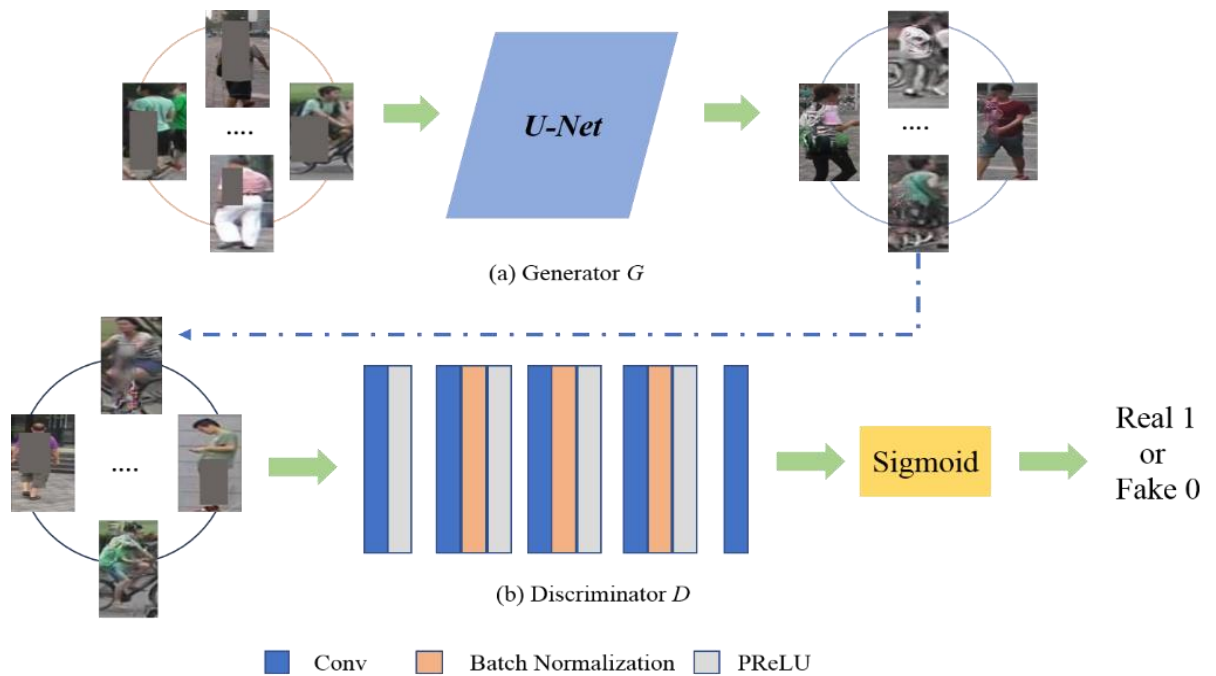

**Figure 2**. The network architecture

***Generative Adversarial Loss.*** In order to force the generated samples have enough ability to puzzle the discriminator *D* and to make the discriminator *D* with strong discrimination. Given an occluded image *I*, the optimization function of GAN model can be written as:

$$\min_G \max_D E_{I \sim P_{data(I)}}[\log(1-D(I,G(I)))] + E_{I \sim P_{data(I,O)}}[\log D(I,O)] \tag{1}$$

in which *O* is the output sample, *D* and *G* represent the discriminator sub-network and generator sub-network, respectively.

***Generator G.*** The generator sub-network aims at learning a non-linear projection function between the occluded image and its corresponding raw image. Through the non-linear projection function, the generator *G* could generate the de-occluded image that similar but not identical to its original images. In this study, we use the U-Net as the generator *G*. It contains a stack of convolutional and de-convolutional layers. The convolutional layers serve as a feature extractor that encodes the basic components of contents for original image while eliminating the occlusion. The de-convolutional decode the encoded contents to restore the content details of raw image. Moreover, several skip connections between the convolutional and de-convolutional are added to preserve more local details.

***Generator loss function.*** Since the generator sub-network tries to generate the de-occluded image that similar but not identical to its corresponding original image, the Euclidean loss is adopted to train the **G** sub-network. Given an occluded image *I*, the loss can be formulated as:

$$L_E(G) = \frac{1}{CMN} \sum_{c=1}^{C} \sum_{m=1}^{M} \sum_{n=1}^{N} \|G(I) - R\|_2^2 \tag{2}$$

in which *N*, *M* and *C* are the height, width and channels of the generated sample, respectively. *R* represent the original image.

***Discriminator D.*** For the discriminator *D*, we use the generated image as negative sample, while using the original image as the positive sample. Then *D* is designed for classifying whether the input person image is a real or fake. Similar to the work of [24], we apply a five layers convolutional network that with PReLU and batch normalization operations for the discriminator. After a set of convolutional layers, the sigmoid layer is followed to output the classification results. The framework of *D* is shown in the bottom of Figure 2.

***Discriminator loss function.*** As descript above, the discriminator *D* is a binary-class network. Given a mixed image set, the objective of *D* can be expressed as:

$$L_D = -\frac{1}{N} \sum_{i=1}^{N} (U_i \log(D(i) - (1-U_i)\log(1-D(i)))) \tag{3}$$

where $U_i$ is the label of *I*, $U_i=0$ represents *i* is a fake sample and $U_i=1$ represents *i* is a real.

## 4. Experiments and Results

In this section, we present the experiment details to testify the effectiveness of the proposed method. The datasets and training details as well as the comparison results are also discussed.

### 4.1 Datasets

Three widely used person ReID datasets are adopted to evaluate the performance of proposed method, including CUHK03 [8], Market-1501 [25] and DukeMTMC-reID [26]. Brief descriptions of the

three datasets are presented as followings:

*CUHK03.* The dataset is one of the largest person ReID datasets which contains 13164 images of 1360 identities. All identities are taken from six camera views, and each pedestrian is captured by two cameras. This data set provides two setting. One automatically annotated by a detector and the other manually annotated by human. We use the labeled set to evaluate our model.

*Market-1501.* This dataset consists of 32643 annotated boxes of 1501 persons. Each pedestrian is collected by at least two cameras and at most six cameras from the front of a supermarket. The boxes of pedestrians are captured by the Deformable Part Model (DPM) detector. The dataset contains 12936 images for training and 19732 images for testing. For this dataset, 12936 images from the training set are used to train the GAN model.

*DukeMTMC-reID.* The DukeMTMC-reID is created for image-based person ReID. It is a subset of DukeMTMC dataset. It contains 1812 identities with 36411 images. These pedestrian images are captured by eight high-resolution surveillance equipment. Among the 1812 identities, 1404 of them captured by more than two camera views and the rest of them are regarded as distractor identifications.

*Evaluation protocols.* The Rank-1, Rank-5 and Rank-10 accuracies are used to quantitatively evaluate the proposed method. The mean Average Precision (mAP) [9] is also adopted for Market-1501 and DukeMTMC-reID datasets. Moreover, we report the re-ranking results based on the *k-reciprocal* encoding [27].

## 4.2 Implementation details

*CNN baselines.* We use the CNN baseline proposed by [28]. The ImageNet pre-trained DenseNet [29] is selected to perform the baseline experiments. The last fully connected (FC) layers of the DenseNets are modified with 1367, 751 and 702 units for CUHK03, Market-1501 and DukeMTMC-reID, respectively. The generated and original images are resized to $256 \times 128$ for training and test. We use the stochastic gradient descent [30] as the optimizer to update the network. For these three datasets, we adopt the same hyper-parameter configurations. We set the dropout probability to 0.5. The momentum and weight decay are set to 0.9 and 5e-4, respectively. We set initial learning rate $\alpha$ to 0.01 and decrease it by dividing 0.1 after each 30 epochs. The total training epoch and mini-batch size are set to 70 and 32, respectively. The runtime of training phase on CUHK03, Market-1501 and DukeMTMC-reID are about 4.5, 5.6 and 5.8 hours, respectively. During the test phase, we use the 2048-dim descriptor produced by the penultimate FC to represent the input image. The relative distance between the probe descriptor and gallery descriptors are calculated before ranking.

*GAN training details.* For the GAN model, we use Pytorch package to implement it. We set the same hyper-parameter configurations for the three datasets. All the occluded and original images are resized into $256 \times 256$ pixels. The Adam [31] is used to optimize the network. We set the momentum and initial learning rate to 0.5 and 0.0002, respectively. We stop the training until 20 epochs. The training processes of CUHK03, Market-1501 and DukeMTMC-reID datasets are roughly converged in 10, 12 and 14 hours, respectively. When test the GAN model, we randomly regenerate the occluded images and input them to the trained GAN model. Then the generated images with its original label information are used for increasing the sample number per person. Several generated images from DukeMTMC-reID dataset are shown as Figure 3. Furthermore, we add 7368, 12936 and 16522 GAN images for CUHK03, Market-1501 and DukeMTMC-reID, respectively.

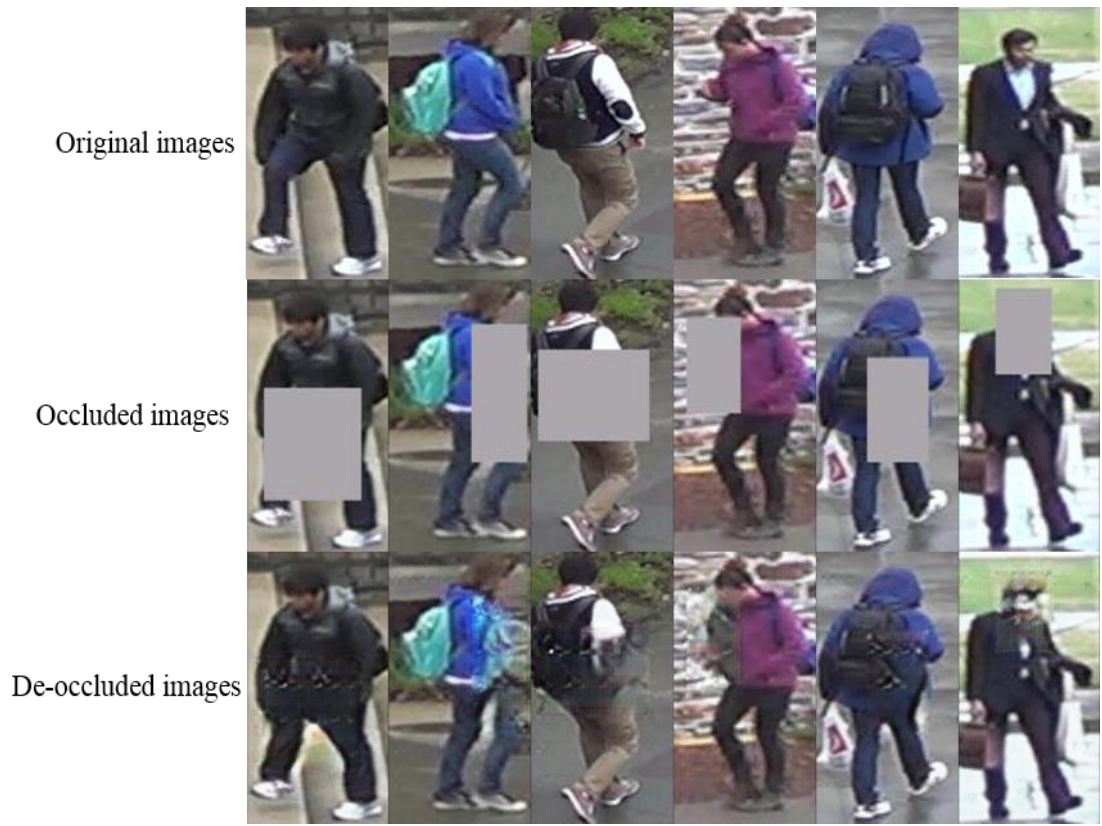

**Figure 3.** Sample generated de-occluded images from the DukeMTMC-reID dataset, from which we can observe that the de-occluded images are similar but not identical to their corresponding original images.

### 4.3 Parameter Sensitivity Test

An important parameter related to our data augmentation approach is the number of GAN images $M$ used for enlarging the training set. We evaluate the influence of different settings of $M$ on the performance of person ReID. The experiments are implemented on Market-1501 under single query mode. Note that the number of training images in Market-1501 is 12936. The performances of baseline with different settings of $M$ on mAP and Rank-1 are shown in Table 1. From Table 1, we can observe that the best performance is achieved when setting $M$ to 12936. When further increasing the number of GAN images, the performance of baseline is degraded. We speculate that this is due to add too many GAN images for training makes the baseline model learn much irrelevant information introduced by the generated noise. Moreover, we find that both the different settings of $M$ consistently improve the performance of baseline model, which demonstrates the effectiveness of our data augmentation approach.

**Table 1.** Experiment results (mAP, Rank1 matching accuracy in %) with different settings of $M$ on the Market-1501 dataset.

|  | Single Query | |
| --- | --- | --- |
| $M$ | mAP | Rank-1 |
| 0 (DesNet-Basl) | 73.58 | 89.67 |
| 12936 | **78.86** | **91.60** |
| 19404 | 78.53 | 91.20 |
| 25872 | 78.12 | 90.90 |
| 32340 | 77.66 | 90.52 |

## 4.4 Comparison with the State-of-the-art Methods

We compare our approach with the following approaches: BoW + KISSME [9], SL [32], DNS [33], Gated Siamese [34], Deep Transfer [35], CAN [20], CNN Embedding [36], SVD-Net [37], HydraPlus-Net [38], CNN+DCGAN [11], TriNet [39], CamStyle [40]. The experimental comparison results are shown in this section below.

*Evaluation on Market-1501.* For the Market-1501, we report both the single-query and multi-query results. The performance on mAP, Rank-1, Rank-5 and Rank-10 of the proposed model are shown in Table2 and Table 3. From Table 2, we can observe that the proposed method achieves rank-1 accuracy=91.80%, mAP = 78.86% with DensNet, using single query mode. When applying the re-ranking, the results further achieve rank-1 accuracy=93.29%, mAP = 90.36%. Besides, compared to the baseline model, our method increases the rank-1 accuracy and mAP by +2.20%, +5.28 % on single query mode, as well as by +4.98%, +2.42% on multi query mode. Both Table1 and Table 2 show that the performance of our model exceeds the compared methods, which demonstrates the effectiveness of the proposed method.

**Table 2.** Results (mAP, Rank1, Rank5 and Rank10 matching accuracy in %) on the Market-1501 dataset in the Single-query. 12936 generated images are added to the training set. '-' means no reported result is available.

|  | Single Query | | | |
|:---:|:---:|:---:|:---:|:---:|
| Method | mAP | Rank-1 | Rank-5 | Rank-10 |
| BoW + KISSME [9] | 20.76 | 44.42 | - | - |
| SL [32] | 26.35 | 51.90 | - | - |
| DNS [33] | 35.68 | 61.02 | - | - |
| Gated Siamese [34] | 39.55 | 65.88 | - | - |
| Deep Transfer [35] | 65.50 | 83.70 | - | - |
| CAN [20] | 35.90 | 60.30 | - | - |
| CNN Embedding [36] | 59.87 | 79.51 | 90.91 | - |
| SVD-Net [37] | 62.10 | 82.3 | 92.3 | 95.2 |
| HydraPlus-Net [38] | - | 76.9 | 91.30 | 94.5 |
| CNN+DCGAN [11] | 56.23 | 78.06 | - | - |
| TriNet [39]+re-rank | 81.07 | 86.67 | 93.38 | - |
| CamStyle [40]+re-rank | 71.55 | 89.49 | - | - |
| DesNet-Basl | 73.58 | 89.67 | 96.61 | 98.16 |
| Ours (DesNet) | 78.86 | 91.60 | **97.34** | **98.46** |
| Our (DesNet)+re-rank | **90.36** | **93.29** | 96.96 | 97.68 |

**Table 3.** Results (mAP, Rank1 and Rank5 matching accuracy in %) on the Market-1501 dataset in the Multi-query. 12936 generated images are added to the training set. '-' means no reported result is available.

|  | Multi Query | | |
| :---: | :---: | :---: | :---: |
| **Method** | **mAP** | **Rank-1** | **Rank-5** |
| BoW + KISSME [9] | 19.47 | 42.64 | - |
| DNS [33] | 46.03 | 71.56 | - |
| Gated Siamese [34] | 48.45 | 76.04 | - |
| Deep Transfer [35] | 73.80 | 89.60 | - |
| CNN Embedding [36] | 70.33 | 85.84 | - |
| CNN+DCGAN [11] | 76.10 | 88.42 | - |
| TriNet [39]+re-rank | 87.18 | 91.75 | 95.78 |
| DesNet-Basl | 79.98 | 91.84 | 98.16 |
| Ours (DesNet) | **84.96** | **94.26** | **97.83** |

*Evaluation on DukeMTMC-reID.* As mentioned above, the 1404 pedestrians are chosen for evaluating the proposed method. We use 16522 images of 702 identities to train the GAN and baseline models. The rest of 702 identities are used for testing the baseline model. From Table 4, we can see that our model achieves rank-1 accuracy=82.04%, mAP = 66.81% with the baseline model. The results are further increased to rank-1 accuracy=86.35%, mAP = 82.81% when applying the re-ranking tool. Moreover, the method increases the rank-1 accuracy and mAP by +2.72%, +2.68 %, respectively.

**Table 4.** Results (mAP, Rank1, Rank5 and Rank10 matching accuracy in %) on the DukeMTMC-reID dataset. 16522 generated images are added to the training set. '-' means no reported result is available.

| **Method** | **mAP** | **Rank-1** | **Rank-5** | **Rank-10** |
| :---: | :---: | :---: | :---: | :---: |
| BoW + KISSME [9] | 12.17 | 25.13 | - | - |
| LOMO+XQDA [41] | 17.04 | 30.75 | - | - |
| CNN+DCGAN [26] | 47.13 | 67.69 | - | - |
| PAN [42] | 51.51 | 71.59 | - | - |
| OIM [43] | 47.40 | 68.10 | - | - |
| CNN embedding [36] | 49.30 | 68.90 | - | - |
| SVD-Net [37] | 56.80 | 76.70 | 86.4 | 89.9 |
| TriNet [39] | 53.50 | 72.44 | - | - |
| ACRN [44] | 51.96 | 72.58 | 84.79 |  |
| DesNet-Basl | 62.89 | 79.36 | 89.73 | 92.41 |
| Ours (DesNet) | 66.81 | 82.04 | 91.24 | 93.89 |
| Ours (DesNet)+re-rank | **82.81** | **86.35** | **92.87** | **94.56** |

*Evaluation on CUHK03.* On CUHK03 dataset, we use the new protocol proposed by [45], in which 767 pedestrians are used for training and the rest of 700 identities are used for test. The evaluation procedure of the new protocol is same as Market-1501. Table 5 shows the comparison results of the proposed model against eight state-of-the-art methods on CUHK03 dataset. From Table 5, we can see that the proposed model achieves 47.31% mAP accuracy and 50.64% rank-1 accuracy, which increases the rank-1 and mAP baseline accuracies by +5.78% and +5.01%, respectively. Table 5 shows that our model outperforms the compared methods, which further confirms the effectiveness of our proposed method.

Table 5. Results (mAP, Rank1 and Rank5 matching accuracy in %) on the CUHK03 dataset (labeled). '-' means no reported results is available. 7368 generated images are added to the training set Under the new evaluation protocol propose by [45]

| Method | mAP | Rank-1 | Rank-5 |
|---|---|---|---|
| BoW + XQDA [9] | 7.29 | 7.93 | - |
| LOMO+XQDA [41] | 13.60 | 14.80 | - |
| PAN [42] +re-rank | 45.80 | 43.90 | 56.86 |
| DPFL [46] | 40.50 | 43.00 | - |
| SVD-Net [37] | 37.83 | 40.93 | - |
| HA-CNN [47] | 41.00 | 44.40 | - |
| HCN+re-rank [48] | 43.00 | 40.90 | - |
| MLFN [49] | 49.20 | 54.70 | - |
| DesNet-Basl | 42.30 | 44.86 | 65.78 |
| Ours(DesNet) | 47.31 | 50.64 | 70.35 |
| Ours(DesNet)+re-rank | **61.95** | **59.78** | **70.64** |

## 5. Conclusions

In this study, a data augmentation method based on generative adversarial network (GAN) was proposed for person ReID. Different from the previous GAN-based methods, we retained the original label information for the generated images. In this way, we increased the sample number per identity. The original images and the generated images composed of the training set to train the baseline model. We conducted the experiment on three large datasets, i.e. CUHK03, Market-1501 and DukeMTMC-reID with the baseline model. The significant improvement on the three datasets demonstrated the effectiveness of the proposed method. Future work will focus on generating GAN images with different degrees of noise disturbing to further enhance the robustness of the baseline model.

## Acknowledgements

This work was supported by the grants of the National Science Foundation of China, Nos. 61472280, 61472173, 61572447, 61732012, 61772370, 61520106006, 31571364, 61672203, 61472282 and 61672382, China Postdoctoral Science Foundation Grant, Nos. 2016M601646 & 2017M611619, and supported by "BAGUI Scholar" Program of Guangxi Province of China. This work has received funding from the European Union's Horizon 2020 research and innovation programme under the Marie klodowska-Curie grant agreement no. 701697.

identification," in *international conference on computer vision*, 2012, pp. 413-422.